\def\year{2021}\relax
\documentclass[letterpaper]{article} 
\usepackage{aaai21}  
\usepackage{times}  
\usepackage{helvet} 
\usepackage{courier}  
\usepackage[hyphens]{url}  
\usepackage{graphicx} 
\usepackage{natbib}  
\usepackage{caption} 
\usepackage{algorithm}
\usepackage{algpseudocode}
\usepackage{subfig}

\title{Deep Innovation Protection: Confronting the Credit Assignment Problem in Training Heterogeneous Neural Architectures}
\author{Sebastian Risi\textsuperscript{\rm 1} and Kenneth O. Stanley\textsuperscript{\rm 2}\\}
\affiliations{
    \textsuperscript{\rm 1}  IT University of Copenhagen, Copenhagen, Denmark, sebr@itu.dk\\ 
    \textsuperscript{\rm 2} Uber AI, San Francisco, CA 94103,  kennethostanley@gmail.com
}
\begin{document}

\maketitle

\begin{abstract}
Deep reinforcement learning approaches have shown impressive results in a variety of different domains, however, more complex heterogeneous architectures such as world models require the different neural components to be trained separately instead of end-to-end. While a simple genetic algorithm recently showed end-to-end training is possible, it failed to solve a more complex 3D task. This paper presents a method called \emph{Deep Innovation Protection} (DIP) that addresses the credit assignment problem in training complex heterogenous neural network models end-to-end for such  environments. The main idea behind the approach is to employ multiobjective optimization to temporally reduce the selection pressure on specific components in multi-component network, allowing other components to adapt. We investigate the emergent representations of these evolved networks, which learn to predict properties important for the survival of the agent, without the need for a specific forward-prediction loss. 
\end{abstract}

\noindent

\section{Introduction}

The ability of the brain to model the world arose from the process of evolution. 
It evolved  because it helped organisms to survive and strive in their particular environments and not because such forward prediction was explicitly optimized for. In contrast to the emergent neural representations in nature, modules of current world model approaches are often directly rewarded for their ability to predict future states of the environment  \citep{schmidhuber1990line, ha2018recurrent,hafner2018learning,wayne2018unsupervised}. While it is undoubtedly useful to be able to explicitly encourage a model to predict what will happen next, here we are interested in the harder problem of agents that should learn to predict what is important for their survival without being explicitly rewarded for it.

A challenge in end-to-end training of complex neural models that does not require each component to be trained separately \citep{ha2018recurrent}, is the well-known credit assignment problem (CAP) \citep{minsky1961steps}. While deep learning has shown to be in general well suited to solve the CAP for deep networks (i.e.\ determining how much each weight contributes to the network's error), evidence suggests that more heterogeneous  networks lack the ``niceness'' of conventional homogeneous networks (see Section~6.1 in \citet{schmidhuber2015deep}), requiring different   training setups for each neural module in combination with evolutionary methods to solve a complex 3D task \citep{ha2018recurrent}. 

To explore this challenge, we are building on the recently introduced world model architecture introduced by \citet{ha2018recurrent} but employ a novel neuroevolutionary optimization method.  This agent model contains three 
different components: \textbf{(1)} a visual module, mapping high-dimensional
inputs to a lower-dimensional representative code, \textbf{(2)} an LSTM-based memory component, and \textbf{(3)} 
a controller component that takes input from the visual and memory module to determine the agent's next action. In the original approach, each component of the world model was trained separately and to perform a different and specialised function, such as predicting the future. While \citet{risi2019deep} demonstrated that these models can also be  trained end-to-end through a population-based genetic algorithm (GA) that exclusively optimizes for final performance, the approach was only applied to the 
simpler  2D car racing domain and it is an open question how such an approach will scale to the more complex CAP in a 3D VizDoom task that first validated the effectiveness of the world model approach. 

Adding support to the hypothesis that CAP is a problem in heterogeneous networks, we show that a simple genetic algorithm fails to find a solution to solving the VizDoom task and ask the question what are the missing ingredients  necessary to encourage the evolution of better performing networks.   
The main insight in this paper is that we can view the optimization of a  heterogeneous neural network (such as world models) as a \emph{co-evolving system of multiple different sub-systems}. The other important CAP insight is that  representational innovations discovered in one subsystem (e.g.\ the visual system learns to track moving objects) require the other sub-systems to adapt. In fact, if the other systems are not given time to adapt, such innovation will likely initially have an adversarial effect on overall performance.

In order to optimize such co-evolving heterogeneous neural systems, we propose to reduce the selection pressure on individuals whose visual or memory system was recently changed, given the controller component time to readapt.  This \emph{Deep Innovation Protection} (DIP) approach is 
able to find a solution to the \texttt{VizDoom:Take Cover} task, which was so far only solved by the original world model approach \citep{ha2018recurrent} and a recent approach based on self-interpretable agents \citep{tang2020neuroevolution}. More interestingly, the emergent models learned to predict events important for the survival of the agent, even though they were not explicitly trained to predict the future. 

Additionally, our investigation into the training process shows that DIP allows  evolution to carefully  orchestrate the training of the  components in these heterogeneous architectures. In other words, DIP is able to successfully \emph{credit} the contributions of the different components to the overall success of the agent. We hope this work inspires more research that focuses on investigating representations emerging from  approaches that do not necessarily only rely on gradient-based optimization.

\section{Deep Innovation Protection}
The hypothesis in this paper is that to  optimize heterogeneous  neural models end-to-end for more complex tasks requires each of its  components to be carefully tuned to work well  together. For example, an innovation in the visual or memory component of the network could adversely impact the controller component, leading to  reduced performance and a complicated CAP. In the long run, such innovation could allow an individual to outperform its predecessors. 

\begin{figure*}
\centering
\includegraphics[width=1.0\textwidth]{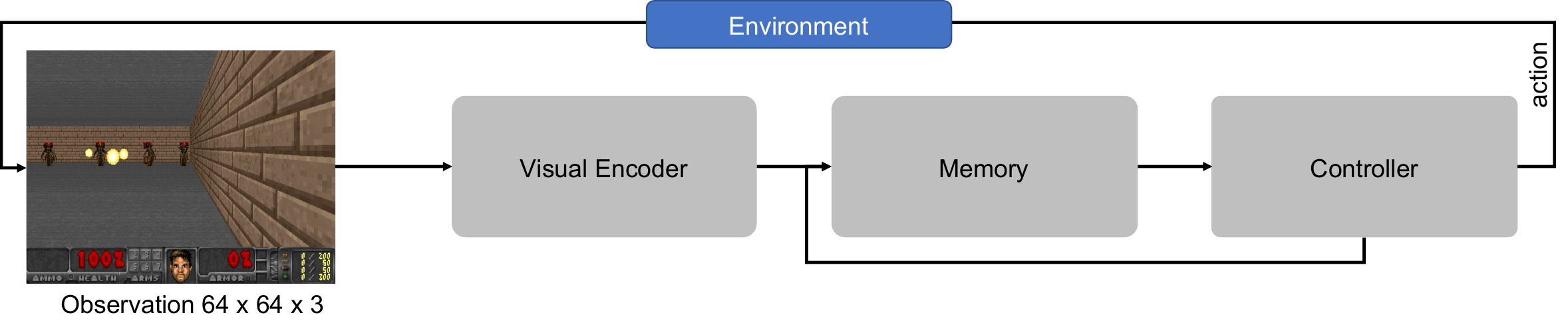} \caption{\emph{Agent Model.} \normalfont The agent model consists of three modules. A visual component that produces a latent code $z_t$ at each time step $t$, which is concatenated with the hidden state $h_t$ of the LSTM-based memory component that takes $z_t$ and previously  performed action $a_{t-1}$ as input. The combined vector ($z_t, h_t$) is input into the controller component to determine the next action of the agent. In this paper, the agent model is trained end-to-end with a multiobjective genetic algorithm.} 
\label{fig:network}
\end{figure*}

The agent's network design is based on the world model network  introduced by \citet{ha2018recurrent}. The network  includes a visual component (VC), implemented as the encoder component of a variational autoencoder that compresses the high-dimensional sensory information into a smaller 32-dimensional representative code (Fig.~\ref{fig:network}). This code is fed into a memory component based on a recurrent LSTM~\citep{hochreiter1997long}, which should predict future representative codes based on previous information. Both the output from the sensory component and the memory component are then fed into a controller that decides on the action the agent should take at each time step.  We train the model end-to-end  with a  genetic algorithm, in which mutations add Gaussian noise to the parameter
vectors of the networks:
$ \theta' = \theta + \sigma \epsilon $, where $	\epsilon \sim N (0, I)$. 

The approach introduced in this paper aims to train  heterogeneous neural systems end-to-end by temporally reducing the selection pressure on individuals with recently changed modules, allowing other components to adapt. For example, in a system in which a mutation  can either affect the visual encoder, MDN-RNN or controller, selection pressure should be reduced if a mutation affects the visual component or MDN-RNN, giving the controller time to readapt to the changes in the learned representation. 
We employ the well-known  multiobjective optimization approach NSGA-II \citep{deb2002fast}, in which a second ``age'' objective keeps track of when a mutation changes either the visual system or the MDN-RNN. Every generation an individual's age is increased by 1, however, if a mutation changes the VC or MDN-RNN, this age objective is set to zero (lower is better). Therefore, if two neural networks reach the same performance (i.e.\ the same final reward), the one that had less time to adapt (i.e.\ whose age is lower) would have a higher chance of being selected for the next generation. The second objective is the accumulated reward received during an  episode. Pseudocode of the approach applied to world models is shown in Algorithm~\ref{alg2}.

It is important to note that this novel approach is different to the traditional usage of ``age'' in multi-objective optimization, in which age is used to increase diversity and keeps track of how long individuals have been in the population  \citep{hornby2006alps,schmidt2011age}. In the approach in this paper, age counts how many generations the controller component of an individual had time to adapt to an unchanged visual and memory system.

\begin{algorithm}
\small
\caption{Deep Innovation Protection}
    \begin{algorithmic}[1]
    \State Generate random population of size N with age objectives set to 0
    \For{$generation=1$\ to\ $i$}
        \For{Individual in Population}
         \State Objective[1] = age
         \State Objective[2] = accumulated task reward
         \State Increase individual's age by 1
         \EndFor
         \State Assign ranks based on Pareto fronts
         \State Generate set of non-dominated solutions
         \State Add solutions, starting from first front, until number solution = N
                 \State Generate child population through binary tournament selection and mutations
                 \State Reset age to 0 for all individuals whose VC or MDN-RNN was mutated
 
    \EndFor
\end{algorithmic}
\label{alg2}
\end{algorithm}

In the original world model approach the visual and memory component were trained separately and through unsupervised learning based on data from random rollouts. We optimize the  multi-component architecture in our work through a  genetic algorithm without evaluating each component individually. In other words, the VC is not directly optimized to reconstruct the original input data and neither is the memory component  optimized to predict the next time step; the whole network is trained in an end-to-end fashion. Here we are interested in what type of neural representations emerge by themselves that allow the agent to solve the given task. 

\section{Experiments}
\begin{figure}
\centering
\subfloat[][\texttt{CarRacing}]
{
\includegraphics[height=0.18\textwidth]{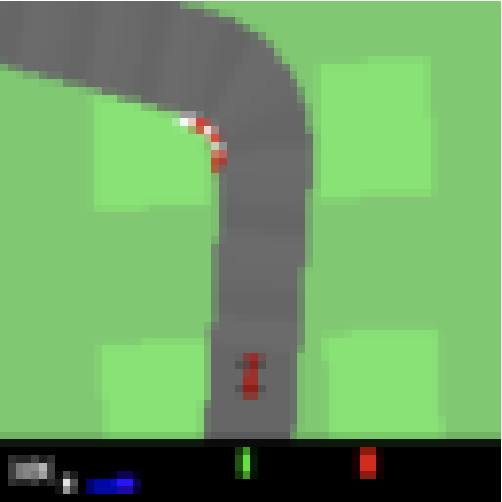}
}
\subfloat[][\texttt{VizDoom}]
{
\includegraphics[height=0.18\textwidth]{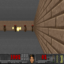}
}
\caption{ In the \texttt{CarRacing-v0} task the agent has to learn to drive across many procedurally generated tracks as fast as possible from 64 $\times$64 RGB color images. In the \texttt{VizDoom:Take Cover} domain the agent has to learn to avoid fireballs and to stay alive as long as possible.}
\label{fig:doom}
\end{figure}

 In the experiments presented here an agent is trained to solve the car racing tasks, and the more challenging VizDoom  task \citep{kempka2016vizdoom} from 64$\times$64 RGB pixel inputs (Fig.~\ref{fig:doom}). These two tasks were chosen to test the generality of the approach, with one requiring 2D top-down control (\texttt{CarRacing-v0}) and the other task requering the control of an agent from a first-person 3D view (VizDoom).
 
 In the continuous control  task \texttt{CarRacing-v0}~\citep{oleg2016} the agent is presented with a new procedurally generated track every episode,  receiving a reward of -0.1 every frame and a reward of +100/$N$ for each visited track tile, where $N$ is the total number of tiles in the track.  The network controlling the agent (Fig.~\ref{fig:network}) has three outputs to control left/right steering, acceleration and braking.  Training agents in procedurally generated environments has shown to significantly increase  their generality  and avoid overfitting \cite{risi2020increasing, justesen2018illuminating, zhang2018study, cobbe2018quantifying}.

 In the \texttt{VizDoom:Take\ Cover} task the agent has to try to stay alive for 2,100 timesteps, while avoiding fireballs shot at it by strafing to the left or the right. The agent receives a +1 reward for every frame it is alive. The network controlling the agent has one output $a$ to control left ($a < -0.3$) and right strafing ($a > 0.3$), or otherwise standing still. In this domain, a solution is defined as surviving for over 750 timesteps, averaged across 100 random rollouts \citep{kempka2016vizdoom}. 

Following the NSGA-II approach, individuals for the next generation are determined stochastically through 2-way tournament selection from the 50\% highest ranked individuals in the population (Algorithm~\ref{alg2}). No crossover operation was employed. The population size was 200.  Because of the randomness in this domain, we evaluate the top three individuals of each generation one additional time to get a better estimate of the true elite.   
We compare a total of four different approaches:
\begin{enumerate}
\item \textbf{Deep innovation protection (DIP):} The age objective is reset to zero when either the VC or MDN-RNN is changed. The idea behind this approach is that the controller should get time to readapt if one of the components that precede it in the network change. 
\item \textbf{Controller innovation  protection:} Here the age objective is set to zero if the controller changes. This setting tests if protecting components upstream can be effective in optimizing heterogeneous neural models.
\item \textbf{MDN-RNN \& Controller innovation protection:}  This setup is the same as the controller protection approach but we additionally reset age if the MDN-RNN changes. On average, this treatment will reset the age objective as often as DIP.
\item \textbf{Random age objective:} In this setup the age objective is assigned a random number between [0, 20] at each evaluation. This treatment tests if better performance can be reached just through introducing more diversity in the population. 
\item \textbf{Standard GA - no innovation protection:} In this non-multi-objective setup, which is the same one as  introduced in \citet{risi2019deep}, only the accumulated reward is taken into account when evaluating individuals.
\end{enumerate}
For all treatments, a mutation has an equal probability to either mutate the visual,
memory, or controller component of the network. Interestingly, while  this approach  performs similarly well to an approach that always mutates all components for the \texttt{CarRacing-v0} task \citep{risi2019deep}, we noticed that it performs significantly worse in the more complicated VizDoom domain. This result suggests that the more complex the tasks, the more important it is to be able to selectively fine-tune  each different  component in a complex neural architecture.


\subsection{Optimization and Model Details}
The genetic algorithm $\sigma$ was determined empirically and set to $0.03$ for the experiments in this paper. The code for the DIP approach is available at: \url{github.com/sebastianrisi/dip}. 

 The sensory model is implemented as a variational autoencoder that compresses the high-dimensional input to a latent vector $z$. The VC takes as input an RGB image of size $64 \times 64 \times 3$, which is passed through four convolutional layers, all with stride 2. 
 The network's weights  are set using the 
 \begin{figure}[t]
    \centering
    \subfloat[][]
{
\includegraphics[height=0.33\textwidth]{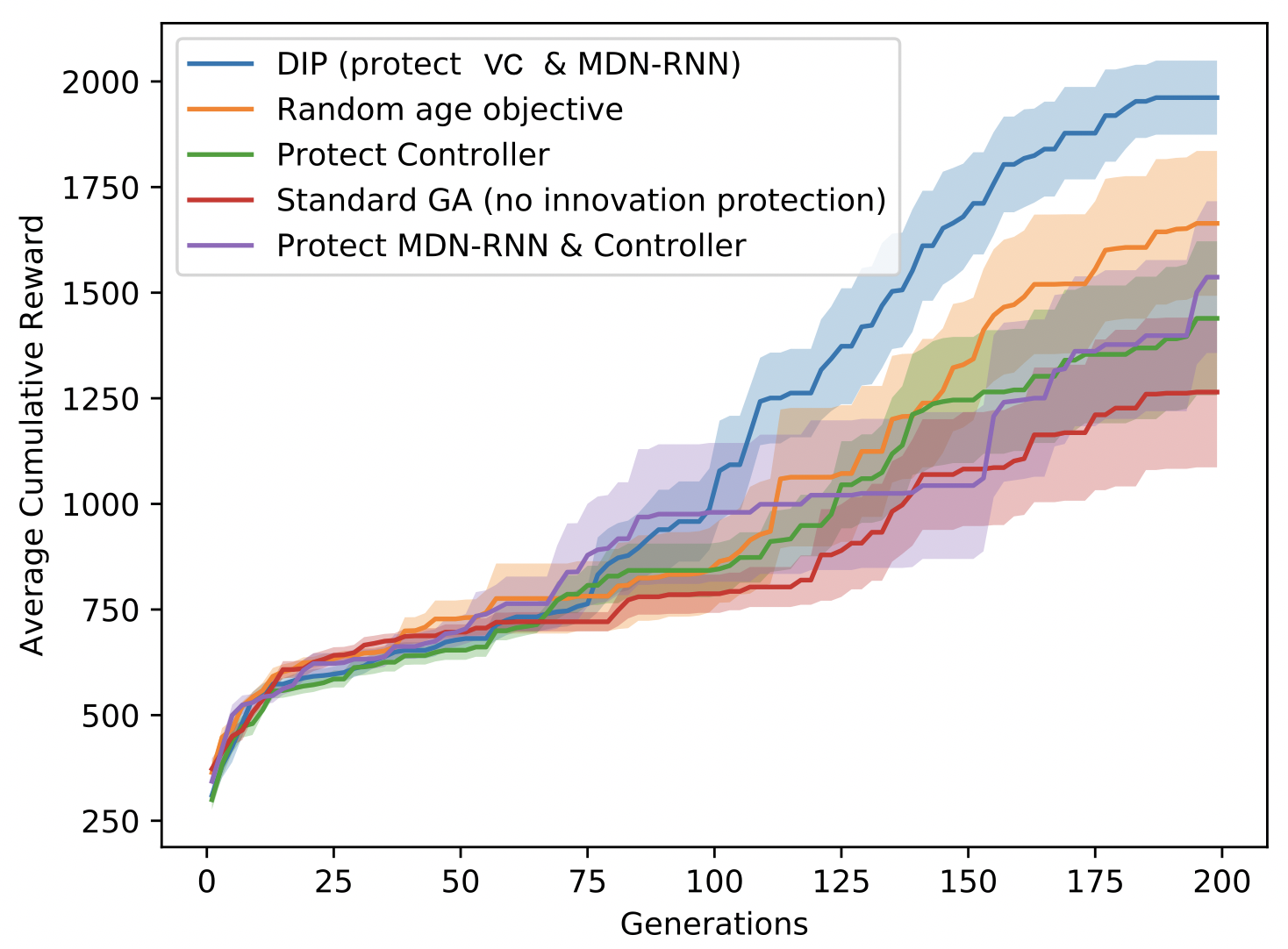}
}
\vspace{-0.05in}
\subfloat[][]
{
\includegraphics[height=0.30\textwidth]{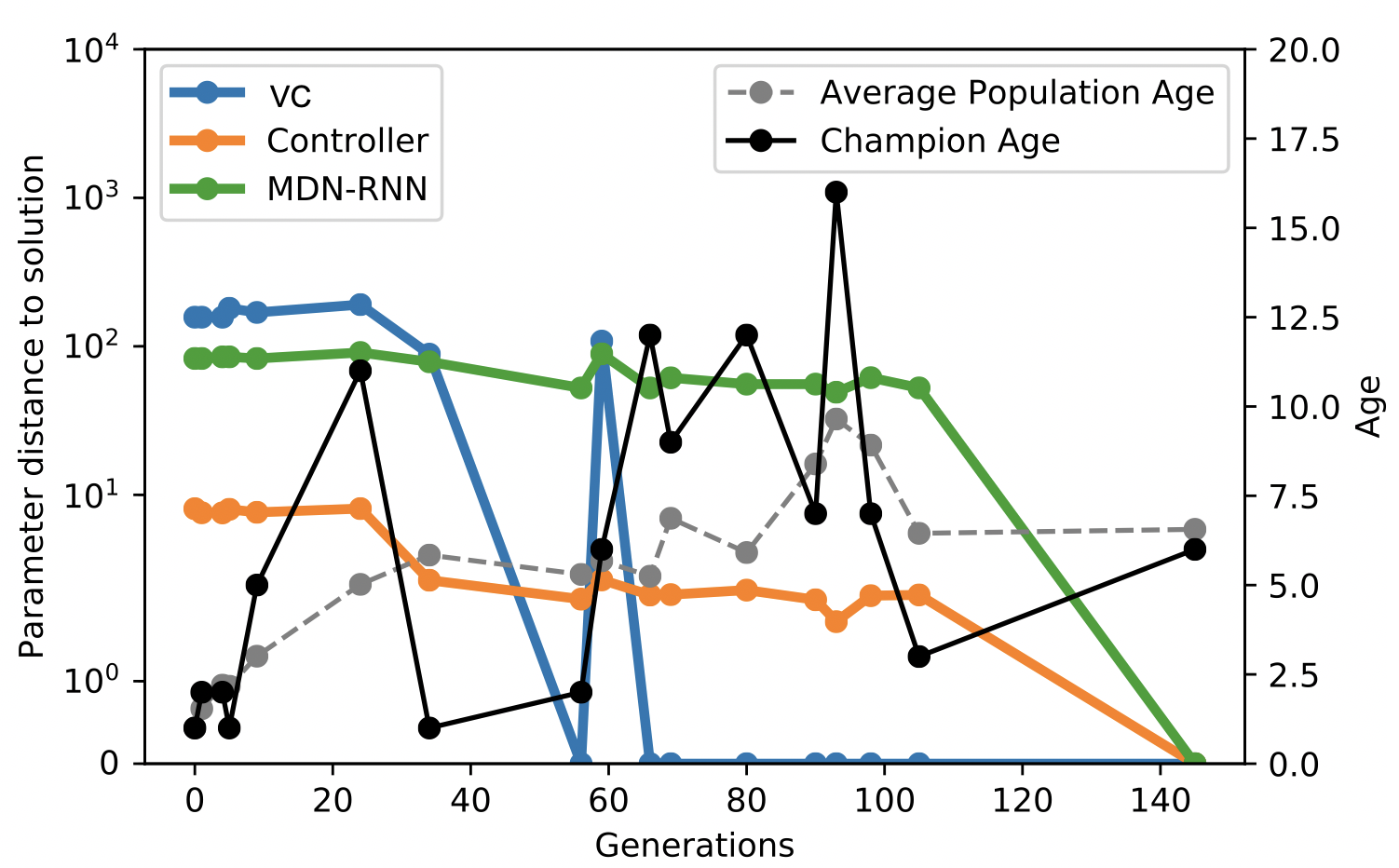}
}
    \caption{ \emph{VizDoom Evolutionary Training.} Shown is (a) mean performance over generations together with one standard error. For one  representative run of  DIP (b), we plot 
     the euclidean distances of the weights of the intermediate solutions (i.e.\ individuals with the highest task  reward discovered so far)  compared to the final solution in addition to their age and the average population age. 
     }
    \vspace{-1.0em}
    \label{fig:training}
\end{figure}
 default He PyTorch initilisation \cite{he2015delving}, with the resulting tensor being  sampled from $\mathcal{U}(-bound, bound)$, where $bound = \sqrt{\frac{1}{fan\_in}}$. The memory model \citep{ha2018recurrent} combines a recurrent LSTM network with a mixture density Gaussian model as network outputs, known as a MDN-RNN~\citep{ha2017neural,graves2013generating}. The network has 256 hidden nodes and models $P(z_{t+1} | a_{t}, z_{t}, h_{t})$, where $a_{t}$ is the action taken by the agent at time $t$ and $h_t$ is the hidden state of the recurrent network. Similar models have previously been used for generating sequences of sketches \citep{ha2017neural} and handwriting \citep{graves2013b}. The controller component is a simple linear model that directly maps   $z_t$ and $h_t$ to actions: $a_t = W_c [z_t h_t] + b_c,$ where $W_c$ and $b_c$ are weight matrix and bias vector. 

\section{Experimental Results}

All results are averaged over ten independent evolutionary runs. 
In the car racing domain we find that there is no noticeable   difference between an approach with and without innovation protection and both can solve the domain with a reward of 905$\pm$80 and 903$\pm$72, respectively.  However, in the more complex VizDoom task (Fig.~\ref{fig:training}a), the DIP approach that protects innovations in both VC and MDN-RNN, significantly outperforms all other approaches during training. The approach is able to find a solution to the task, effectively avoiding fireballs and reaching  an average score of 824.33 (sd $\pm$ 491.59). 

\begin{figure}
    \centering
    \includegraphics[width=1.0\columnwidth]{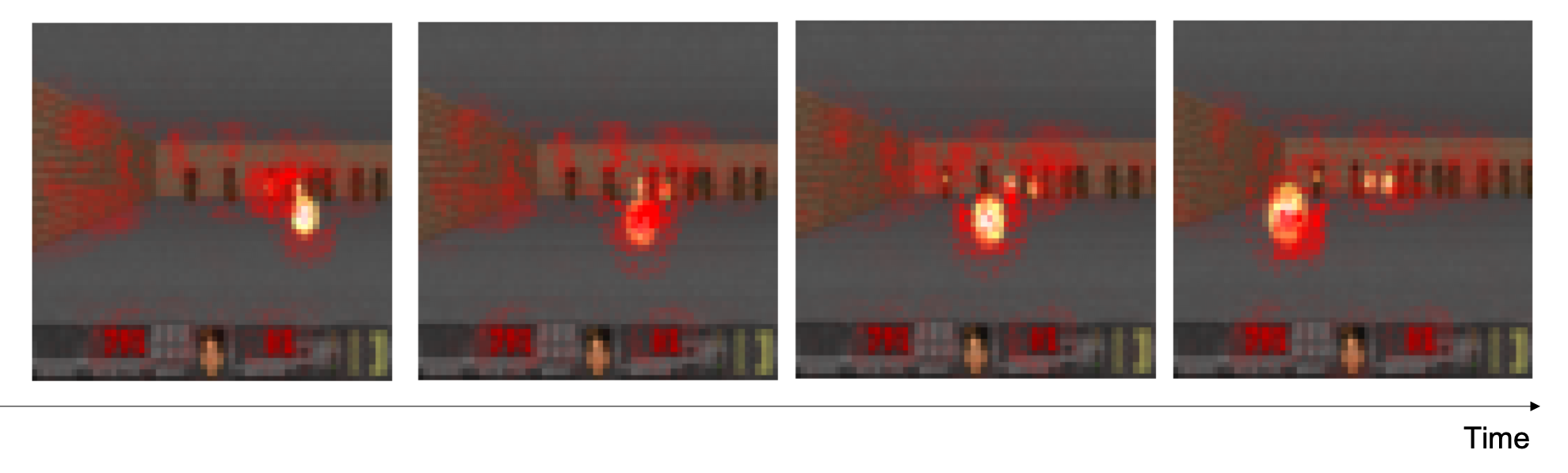}
    \caption{ \emph{Still frames of a learned policy.} The agent learned to primarily pay attention to the walls and fireballs, while ignoring the floor and ceiling. Interestingly the agent also seems to pay attention to the health and ammo indicator. %
    }
    \label{fig:visual_rep}
    \vspace{-1.0em}
\end{figure}

To better understand the network's behavior, we calculate perturbation-based saliency maps to determine the parts of the environment the agent is paying attention to (Fig.~\ref{fig:visual_rep}). The idea behind perturbation-based saliency maps is to measure to what extent the output of the model changes if parts of the input image are altered \citep{greydanus2017visualizing}.  Not surprisingly, the agent learned to pay particular attention to the walls, fireballs, and  the position of the monsters. 

The better performance of the random age objective compared to no innovation protection  suggests that increasing  diversity in the population improves performance but less effectively than selectivity resetting  age  as in DIP. Interestingly, the controller and the MDN-RNN\&Controller protection approach perform less well, confirming our hypothesis that it is important to protect innovations upstream in the network for downstream components. 

\begin{figure}[t]
\centering
\subfloat[][z+hidden]
{
\includegraphics[height=0.09\textwidth]{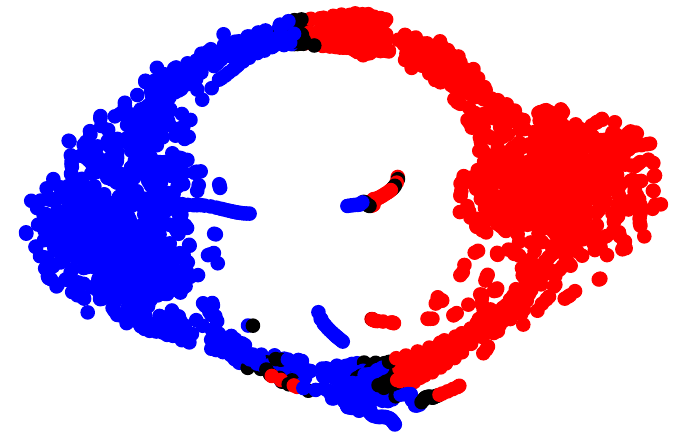}
}
\subfloat[][z alone]
{
\includegraphics[height=0.09\textwidth]{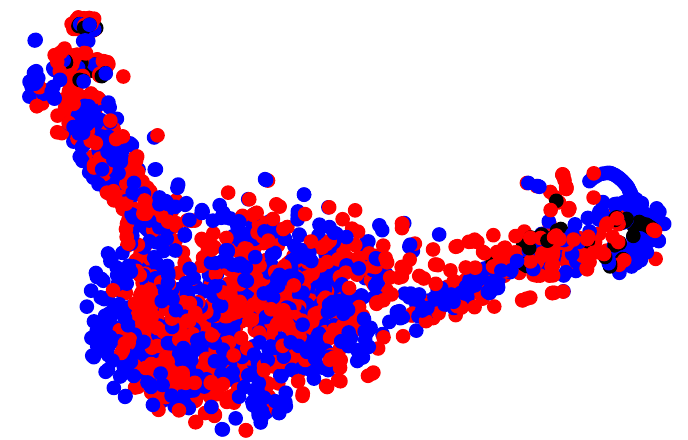}
}
\subfloat[][hidden alone]
{
\includegraphics[height=0.09\textwidth]{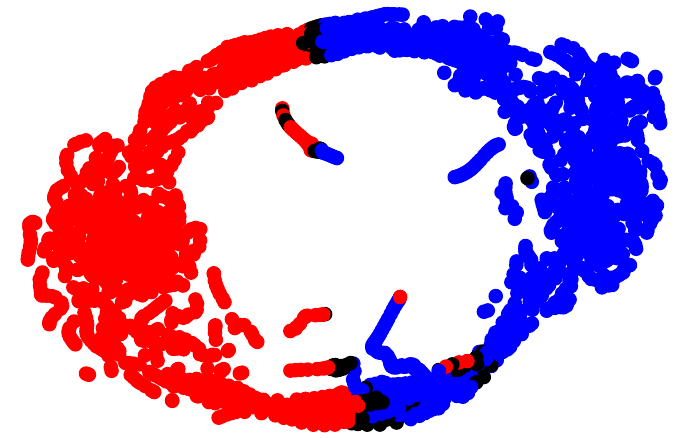}
}
\caption{t-SNE mapping of the latent+hidden vector (a), latent vector alone (b), and hidden vector alone (c). While the compressed latent vector is not enough to infer the correct action (b), the hidden LSTM vector alone contains enough information for the agent to decide on the correct action (c). Red = strafe left, blue =  strafe right, black = no movement.}
\label{fig:tsne}
\end{figure}

\textbf{Learned Representations } We further investigate what type of world model can emerge  from an evolutionary process that does not directly optimize
for forward prediction or reconstruction loss. To gain insights into the  learned representations we employ the t-SNE dimensionality reduction technique \citep{maaten2008visualizing}, which has  proven valuable for visualizing the inner workings of deep neural networks \citep{such2018atari,mnih2015human}. We are particularly interested in the information contained in the compressed 32-dimensional vector of the VC and the information stored in the hidden states of the MDN-RNN (which are both fed into the controller that decides on the agent's action). Different combinations of sequences of these latent vectors collected during one rollout are visualized in two dimensions in Fig.~\ref{fig:tsne}. Interestingly, while the 32-dimensional $z$ vector from the VC does not contain enough information to infer the correct action, either the hidden state alone or in combination with $z$ results in grouping the states into two distinct classes (one for moving left and one for moving right).  The temporal dimension captured by the recurrent network proves  invaluable in deciding what action is best. For example, not getting stuck in a position that makes avoiding incoming fireballs impossible, seems to require a level of forward prediction by the agent. To gain a deeper understanding of this issue we look more closely into the learned temporal representation next.



\begin{figure*}[t]
\centering
\subfloat[][]
{
\includegraphics[width=1.0\textwidth]{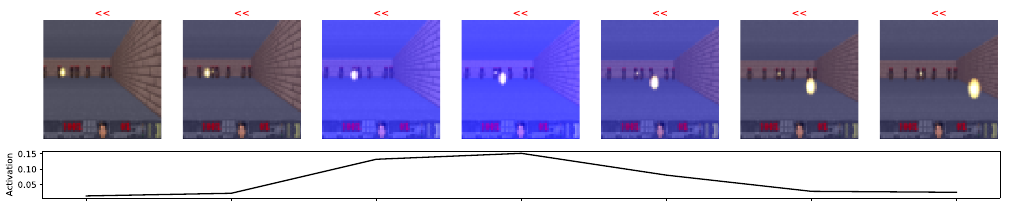}
}
\\
\subfloat[][]
{
\includegraphics[width=1.0\textwidth]{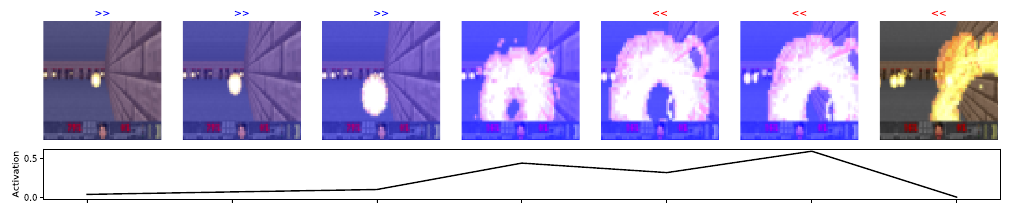}
}
\caption{\emph{Average activation levels of LSTM in two different situations.} For visualization purposes only, images are colored more or less blue depending on the LSTM activations.  The  forward model seems to have learned to predict if a fireball would hit the agent at the current position. In (a) the agent can take advantage of that information to avoid the fireball while the agent does not have enough time to escape in situation (b) and gets hit. Shown on top are the actions the agent takes in each frame.}
\label{fig:lstm_dynamics}
\end{figure*}

\textbf{Learned Forward Model Dynamics }
\label{sec:temporal}
In order to analyze the learned temporal dynamics of the forward model, we are taking a closer look at the average activation $x_t$ of all 256 hidden nodes at time step $t$ and how much they differ from the overall average across all time steps $\bar{X}=  \frac{1}{N} \sum_1^N \bar{x_t}$. The variance of $\bar{x_t}$ is thus calculated as $\sigma_t = (\bar{X}-\bar{x_t})^2 $, and normalized to the range $[0, 1]$ before plotting. The hypothesis is that activation levels far from the mean might indicate a higher importance and should have a greater impact on the agent's controller component. In other words, they likely indicate critical situations in which the agent needs to pay particular attention to the predictions of the MDN-RNN. Fig.~\ref{fig:lstm_dynamics} depicts frames from the learned policies in two different situations, which shows  that the magnitude of LSTM activations are closely tied to specific situations. The forward model does not seem to  react to fireballs by themselves but instead depends on the agent being in the line of impact of an approaching fireball, which is critical information for the agent to stay alive. 

\begin{figure}[t]
    \centering
    \includegraphics[width=1.0\columnwidth]{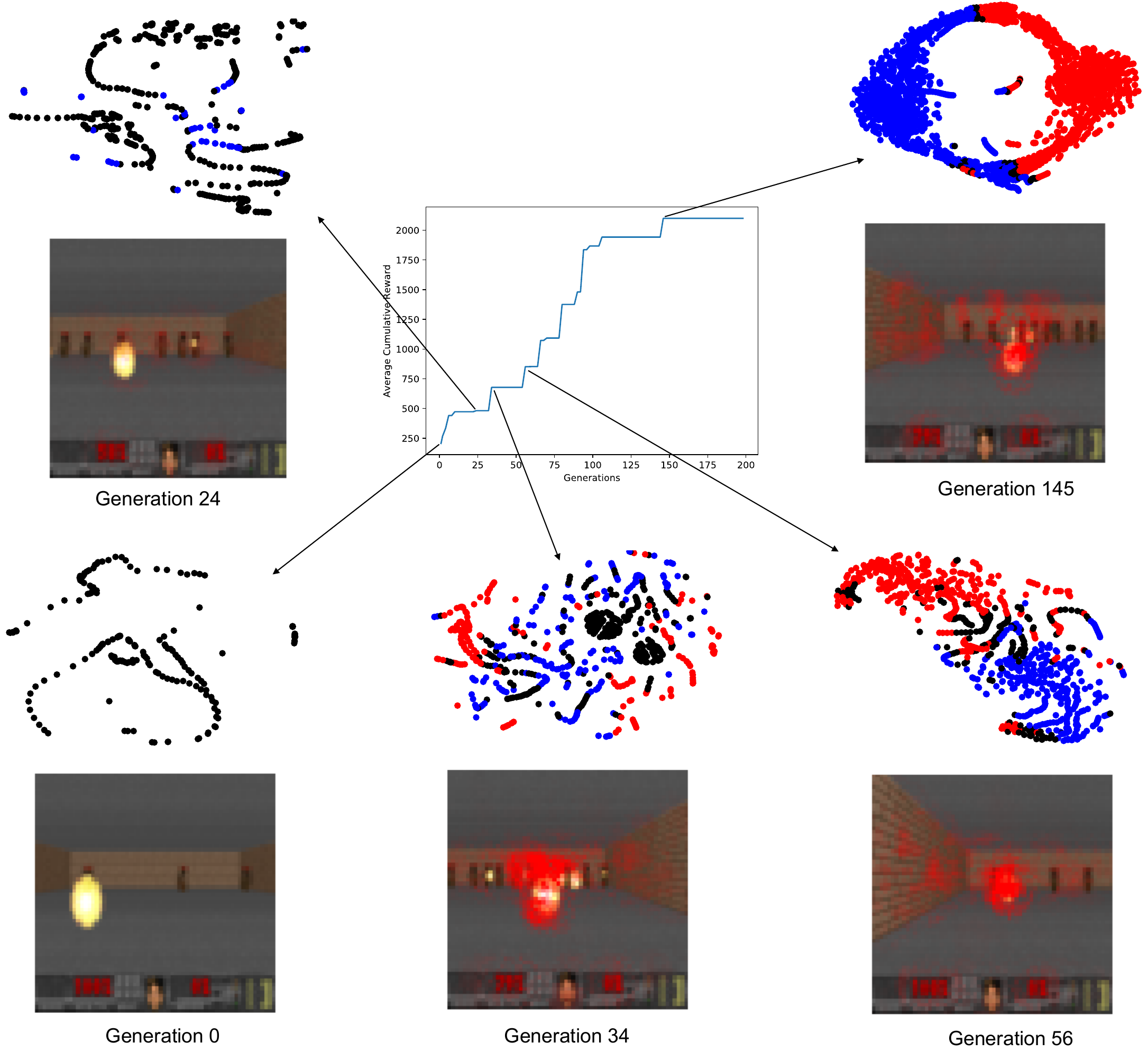}
    \caption{\emph{Development of the evolved representation.} Shown are t-SNE mappings of the 288-dimensional vectors (32-dimensional latent vectors + 256-dimensional hidden state vector) together with saliency maps of specific game situations. Early on in evolution the agent starts paying attention to the fireballs (generation 24) but only moves to the right (blue) or stands still (black). Starting around generation 34 the agent starts to move to the left and right, with the saliency maps becoming more pronounced. From generation 56 on the compressed learned representation (latent vector+hidden state vector) allows the agent to infer the correct action almost all the time. The champion discovered in generation 145 discovered a visual encoder and LSTM mapping that shows a clear division for left and right strafing actions.}
    \label{fig:dev}
    \vspace{-1.0em}
\end{figure}

\textbf{Evolutionary Innovations } In addition to analyzing the  learned representations of the final networks, it is interesting to study the  different stepping stones evolution discovered to solve the VizDoom task. We show one particular evolutionary run in  Fig.~\ref{fig:dev}, with other ones following similar progressions. In the first 30 generations the agent starts to learn to  pay attention to fireballs but only tries avoiding them by either standing still or moving to the right. A jump in performance happens around generation 34 when the agent starts to discover moving to either the left or right; however, the learned representation between moving left or right is not well defined yet. This changes around generation 56, leading to another jump in fitness and some generations of quick fine-tuning later  the agent is able to differentiate well between situations requiring different actions, managing to survive for the whole length of the episode. Motivated by the approach of \citet{raghu2017svcca} to analyse the gradient descent-based training of neural networks, we investigate the weight distances of the  world model components of the best-performing networks found during training to the final solution representation (Fig.~\ref{fig:training}b). The VC is the  component with the steepest decrease in distance with a noticeable jump around generation 60 due to another lineage taking over. The MDN-RNN is optimized slowest, which is likely due to the fact that the correct forward model dynamics are more complicated to discover than the visual component. These results suggest that DIP is able to orchestrate the training of these heterogeneous world model architectures in an automated way, successfully solving the underlying CAP. 

\begin{figure}
\centering
\includegraphics[height=0.25\textwidth]{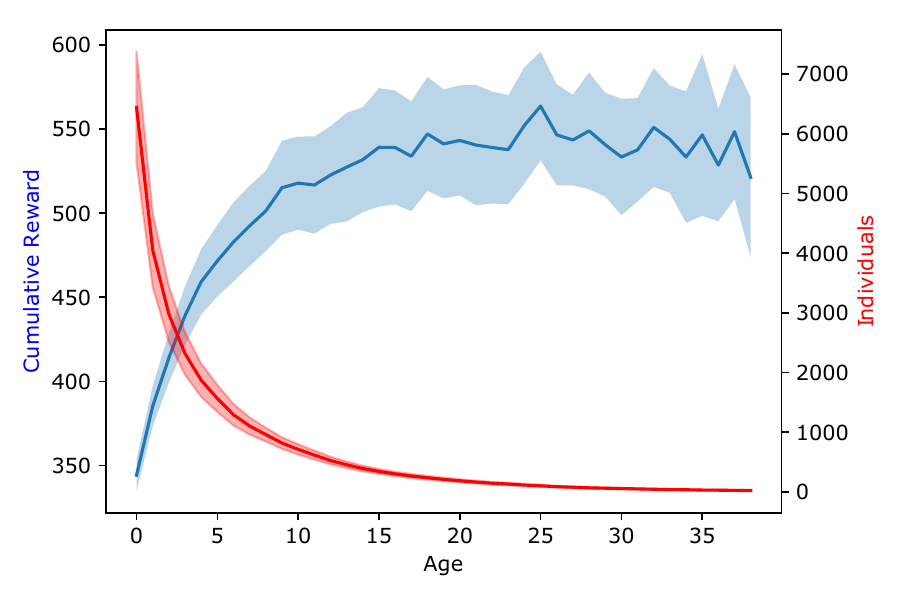}
\caption{Average reward across ages and number of individuals per age.}
\label{fig:re_age}
\end{figure}

\textbf{Reward and Age Objective } We performed an  analysis of the  (1) cumulative reward per age and (2) the number of individuals with a certain age  averaged across all ten runs and all generations   (Fig.~\ref{fig:re_age}). While the average reward increases with age, there are fewer and fewer individuals at higher age levels. This result suggest that the two objectives 
are in competition with each other, motivating the choice for a multi-objective optimization approach; 
staying alive for longer becomes increasingly difficult and a high age needs to be compensated for by a high task reward. 

\textbf{Sample Efficiency Comparison } While evolutionary algorithms are typically regarded as requiring many samples, DIP is surprisingly sample efficient and competitive with other solutions to the DoomTakeCover and CarRacing task. The other reported solutions that solve both of these tasks are the  world model approach by \citet{ha2018recurrent} and an evolutionary self-attention approach \cite{tang2020neuroevolution}. In case of the CarRacing task, \citet{tang2020neuroevolution} report that they can solve the tasks reliable after 1,000 generations (with a slightly larger population size of 256 compared to our population size of 200).  The world model approach uses a mix of different methods \citep{ha2018recurrent}, which makes comparing sample efficiency slightly more complicated. The world model approach finds a solution to the CarRacing task in 1,800 generations with an already trained VAE and MDN-RNN. DIP can solve CarRacing after 1,200 generations (without requiring pre-training) and is thus similarly sample efficient to the end-to-end training approach in \citet{tang2020neuroevolution}. The purely evolutionary training in \citet{tang2020neuroevolution} can reliable solve the DoomTakeCover task after around 1,000 generations. DIP solves the tasks in only 200 generations. The world model approach only trains the DoomTakeCover agent in a simulated dream environment and then transfers the controller to the actual environment. Evolutionary training of the learned world model is fast, since it doesn't require simulated graphics, and takes around 1,500 generations. However, it relies on training the VAE and MDN-RNN with 10,000 random rollouts.

\section{Related Work}

A variety of different RL algorithms have recently been shown to work well on a diverse set of problems when combined with the representative power of deep neural networks \citep{mnih2015human,schulman2015trust,schulman2017proximal}. While most approaches are based on variations of Q-learning \cite{mnih2015human} or policy gradient methods \citep{schulman2015trust,schulman2017proximal}, recently evolutionary-based methods have emerged as a promising  alternative for some domains  \citep{such2017deep,salimans2017evolution}. 
\citet{salimans2017evolution} showed that a type of evolution strategy (ES) can reach competitive performance in the Atari benchmark and at controlling robots in MuJoCo. Additionally, \citet{such2017deep} demonstrated that a simple genetic algorithm is in fact able to reach similar performance to  deep RL methods such as DQN or A3C.  Earlier approaches that evolved neural networks for RL tasks worked well in complex RL tasks with lower-dimensional input spaces \citep{stanley2002evolving,floreano2008neuroevolution,risi2017neuroevolution}. Evolutionary approaches solving 3D tasks directly from  pixels has so far proven difficult although a few notable approaches exist \cite{koutnik2013evolving,alvernaz2017autoencoder,poulsen2017dlne,lehman2018safe}. 


For complex agent models,  different network components can be trained separately \citep{wahlstrom2015pixels, ha2018recurrent}. For example, in the world model approach \citep{ha2018recurrent}, the authors first train a variational autoencoder (VAE) on 10,000 rollouts from a random policy to compress the high-dimensional sensory data and then train a recurrent network to predict the next latent code. Only after this process is a smaller controller network trained to perform the actual task, taking information from both the VAE and recurrent network as input to determine the action the agent should perform. 

Evolutionary approaches solving 3D tasks directly from  pixels has so far proven difficult although a few notable approaches exist. \citet{koutnik2013evolving} evolved an indirectly encoded and recurrent controller for car driving in TORCS, which learned to drive based on a raw 64$\times$64 pixel image. The approach was based on an indirect encoding of the network's weights analogous to the JPEG compression in images. To scale to 3D FPS tasks,  \citet{alvernaz2017autoencoder} first  trained an autoencoder in an unsupervised way  and then evolved the controller giving the compressed representation as input. In another approach,  \citet{poulsen2017dlne}  trained an object recognizer in a supervised way  and then in a separate step evolved a controller module. More recently, \citet{lehman2018safe} introduced an approach called \emph{safe mutations}, in which the magnitude of mutations to weight connections is scaled based on the sensitivity of the network's output to that weight. It allowed the evolution of large-scale deep networks  for a  simple 3D maze task and is a complementary approach that could be combined with  DIP. 

The approach introduced in this paper can be viewed as a form of diversity maintenance, in which selection pressure on certain mutated neural networks is reduced. Many other methods for encouraging diversity \citep{mouret2012encouraging} were invented by the evolutionary computation community, such as novelty search \citep{lehman2008exploiting}, quality diversity \citep{pugh2016quality}, or speciation \citep{stanley2002evolving}. 

For increasing diversity, algorithms often introduce new individuals into the population. In the ALPS approach by \citet{hornby2006alps}, the population is segregated into different  layers depending on when they were introduced into the population and newly generated individuals are introduced into the "newest" layer to increase diversity. \citet{schmidt2011age} combine this idea with a multi-objective approach, in which individuals are rewarded for performance and for how many generations have passed since they have been introduced into the population. Similar to the approach by  \citet{Cheney20170937} to co-evolve morphologies and neural controller, and in contrast to previous approaches \citep{hornby2006alps,schmidt2011age}, DIP does not introduce new random individuals into the generation but instead resets the ``age'' of individuals whose sensory or memory system have been mutated. That is, it is not a measure of how long the individual has been in the population. 

Approaches to learning dynamical models have mainly focused on gradient descent-based methods, with early work on RNNs in the 1990s \citep{schmidhuber1990line}. More recent work includes PILCO \citep{deisenroth2011pilco}, which is a probabilistic model-based policy search method and Black-DROPS \citep{chatzilygeroudis2017black} that employs CMA-ES for data-efficient optimization of complex control problems. Additionally, interest has increased in learning  dynamical models directly from high-dimensional images for robotic tasks \citep{watter2015embed, hafner2018learning}  and also video games \citep{guzdial2017game}. Work on evolving forward models has mainly focused on neural networks that contain orders of magnitude fewer connections and lower-dimensional feature vectors   \citep{norouzzadeh2016neuromodulation} than the  models in this paper. 

\section{Discussion and Future Work}
The paper demonstrated that a predictive  representation for a 3D task can emerge under the right circumstances without being explicitly rewarded for it. To encourage this  emergence and address the inherent credit assignment problem of complex heterogeneous networks, we introduced the \emph{deep innovation protection} approach that can dynamically reduce the selection pressure for different components in  such neural architectures. The main insight is that when components upstream in the  network change, such as the visual or memory system in a world model, components downstream need time to adapt to  changes in those learned representations.   

The neural model learned to represent situations that require similar actions with similar latent and hidden codes (Fig.~\ref{fig:tsne} and \ref{fig:dev}).  Additionally, without a specific forward-prediction loss, the agent learned to predict ``useful'' events that are necessary for its survival  (e.g.\ predicting when the agent is in the line-of-fire of a fireball).  In the future it will be interesting to compare the differences and similarities of  emergent representations and learning dynamics resulting from evolutionary and gradient descent-based optimization approaches \citep{raghu2017svcca}.



A natural extension to this work is to evolve the neural architectures in addition to the weights of the network. Searching for neural architectures in RL has previously only  been applied to smaller networks \citep{risi2012enhanced,stanley2002evolving,stanley2019designing, gaier2019weight,risi2017neuroevolution,floreano2008neuroevolution} but could potentially now be scaled to more complex tasks. While our innovation protection approach is based on evolution, ideas presented here could also be incorporated in gradient descent-based approaches that optimize neural systems with multiple interacting components end-to-end.

\section{Broader Impact}
The ethical and future societal consequences of this work are hard to predict but likely similar to other work dealing with solving complex reinforcement learning problems. Because these approaches are rather task agnostic, they could potentially be used to train autonomous robots or drones in areas that have both a positive and negative impact on society. While positive application can include delivery drones than can learn from visual feedback to reach otherwise hard to access places, other more worrisome military applications are also imaginable. The approach presented in this paper is far from being deployed in these areas, but it its important to discuss its potential long-term consequences early on.

\section{Acknowledgments}
We would like to thank Nick Cheney, Jeff Clune, David Ha, and the members of Uber AI for helpful comments and discussions on ideas in this paper.

\bibliography{main}

\end{document}